\documentclass{article}
\usepackage{ijcai26}

\usepackage{times}
\usepackage{soul}
\usepackage{url}
\usepackage[hidelinks]{hyperref}
\usepackage[utf8]{inputenc}
\usepackage{newunicodechar}
\DeclareUnicodeCharacter{0301}{\'}
\newunicodechar{̀}{\`{}}
\usepackage{tipa}
\newunicodechar{ə}{\textschwa}
\usepackage[small]{caption}
\usepackage{graphicx}
\usepackage{amsmath}
\usepackage{amsthm}
\usepackage{booktabs}
\usepackage{algorithm}
\usepackage{algorithmic}
\usepackage[switch]{lineno}
\usepackage{graphicx}
\usepackage{subcaption}
\graphicspath{{images/}}

\title{Embedding Initialization for Unseen Low-resource Languages in Multilingual NMT: A Case Study on Limbum–English Translation}

\author{
Samiratu Ntohsi
\and
Neza David Tuyishimire
\and
Anesu Kafesu
\and
Marvin Ogore
\and
Samuel Oluwajunwonlo Babalola
\and
Oche Ankeli
\affiliations
Ruzivo Research Lab, African Leadership University
\emails
\{sntohsi, dtuyishimire, mogore\}@alueducation.com,
\{a.kafesu, s.babalola, o.ankeli\}@alustudent.com
}
\begin{document}
\maketitle

\begin{abstract}
Multilingual neural machine translation models such as NLLB-200 cover 200
languages but leave thousands unsupported, including most Grassfields Bantu
languages of Cameroon. When fine-tuning these models for an unseen language, practitioners must choose a proxy language token, yet no principled method
exists for this selection. We implemented an embedding initialization strategy where a language token is the average of embeddings from multiple typologically related languages already in the model. We evaluate this approach on Limbum-to-English translation using a parallel corpus of 8,837 sentence pairs from New Testament text and a bilingual dictionary. We compare models: NLLB-200 zero-shot (chrF2++ = 12.5), a Transformer trained from scratch (chrF2++ = 14.5), NLLB-200, fine-tuned with a Swahili proxy token (chrF2++ = 47.3), and NLLB-200 with our averaged embedding initialization (chrF2++ = 46.7). We find that the multi-language initialization achieves performance comparable to the best single-language proxy. Both NLLB-200 variants improve over the from-scratch baseline by over 32 chrF2++ points. These results show that multilingual transfer is the dominant factor in extremely low-resource Bantu translation while eliminating the need for heuristic proxy selection. However, all systems fail to preserve tonal diacritics, highlighting an open challenge. We make our dataset and code available to support further research.
\end{abstract}

\section{Introduction}

NLLB-200, the largest open multilingual translation model, supports 200
languages \cite{costa2022no}. This is a substantial achievement, yet
it covers fewer than 3\% of the world's approximately 7,000 living languages \cite{eberhard2020ethnologue}. The gap is more prominent for sub-Saharan Africa, where
over 2,000 languages are spoken, but only a fraction appear in any NLP system
\cite{joshi2020state}. For speakers of these unsupported languages, the
benefits of modern NMT remain inaccessible.

When researchers attempt to fine-tune a multilingual model for an unseen
language, they face a practical problem: the model has no language token for
the new language. The standard workaround is to repurpose the token of a
related language as a proxy. \cite{adelani2022few}, for instance, map
several African languages to typologically similar proxies already in the
model. This approach works, but it depends on language-specific knowledge
about which proxy to choose, and no systematic comparison of proxy selection
strategies has been conducted.

We experiment with an approach where, rather than selecting a single proxy, we
initialize the embedding of a new language token by averaging the embeddings
of $k$ related languages already in the NLLB model. The intuition is that if multiple related languages occupy a region of the model's embedding space, their centroid should capture features common to the language family without committing to any single member. The method requires no architecture modification beyond adding one token to the
vocabulary and resizing the embedding matrix.

We evaluate this approach on Limbum (ISO 639-3: `lmp`), a Grassfields Bantu language of Cameroon with no prior computational resources, making it a useful test case for extremely low-resource translation: no treebank, no word embeddings, no parallel corpus, and no entry in NLLB-200. We assembled the first Limbum–English parallel corpus (8,837 sentence pairs) from New
Testament text and a bilingual dictionary, and we compared four models in a stepped ablation:
\begin{itemize}
    \item \textbf{S0}: NLLB-200 zero-shot (no fine-tuning)
    \item \textbf{S1}: A Transformer trained from scratch on our corpus
    \item \textbf{S2a}: NLLB-200 fine-tuned using Swahili (\texttt{swh\_Latn}) as a single-language proxy
    \item \textbf{S2b}: NLLB-200 fine-tuned with a new (\texttt{`lmp\_Latn`}) token initialized by averaging three Bantu language embeddings
\end{itemize}
The results show that S2b achieves chrF2++ of 46.7, comparable to S2a at
47.3 (overlapping 95\% confidence intervals). The primary finding is the gap
between S1 (14.5) and S2a/S2b ($\approx 47$): multilingual transfer accounts for a
32-point improvement, while the choice of initialization strategy accounts
for less than one point. This suggests that for researchers working with
unseen Bantu languages, the averaged initialization is a safe default that
avoids the proxy selection problem entirely.

Our contributions, in order of significance, are as follows.

\begin{enumerate}
    \item \textbf{The first parallel corpus for Limbum-English} (8,837 pairs from two domains), released publicly to support further Limbum NLP research.
    
    \item \textbf{An empirical demonstration} that fine-tuning NLLB-200 yields chrF2++ $\approx 47$ for a genuinely unseen Bantu language, using only 7,070 training pairs on Google Colab Pro T4 GPU.
    
    \item \textbf{An embedding initialization method} for unseen languages that matches single-proxy fine-tuning while eliminating the proxy selection problem.
    
    \item \textbf{Ablation analyses}, a learning curve over four data fractions, and an NT-only ablation quantifying data efficiency and the contribution of supplementary lexical resources.
    
    \item \textbf{A documented failure case} in tonal diacritic preservation, highlighting an open challenge for NMT systems applied to tonal African languages.
\end{enumerate}

\section{Background}

\subsection{Multilingual NMT and Language Tokens}
Modern multilingual NMT systems prepend a language token to each source
sentence to signal the desired translation direction \cite{johnson2017google}.
In NLLB-200, these tokens follow the format \texttt{\{iso\}\_\{script\}} (e.g., \texttt{swh\_Latn} for Latin-script Swahili). Each token has a learned embedding
that encodes language identity. During fine-tuning, this embedding adapts
to represent the characteristics of the source language in the model's
shared multilingual space.

NLLB-200 \cite{costa2022no} is a 600M\-parameter encoder-decoder
model trained on parallel data spanning 200 languages, with particular
attention to low-resource African and Asian languages. Its SentencePiece
vocabulary of 256,000 tokens provides subword coverage for all supported
languages, including Unicode combining characters used in tonal
orthographies. For languages outside this set, however, there is no
established adaptation procedure.

\subsection{Adapting to Unseen Languages}

Transfer learning from a high-resource "parent" model to a low-resource
"child" language has been shown to improve NMT quality when direct training
data is scarce \cite{zoph2016transfer}. \cite{neubig2018rapid} demonstrated rapid
adaptation to new languages by fine-tuning multilingual models on small
parallel corpora, finding that related source languages provide better
initialization than unrelated ones. Adelani et al. (2022) showed that as
few as 3,000 parallel sentences suffice for meaningful fine-tuning of
pretrained multilingual models for African language translation.

Within the African NLP community, several efforts have demonstrated the
viability of multilingual transfer for under-resourced languages. \cite{emezue2021mmtafrica} developed MMTAfrica, a multilingual translation system
covering six African languages, showing that multi-way African language
translation benefits from shared multilingual representations. \cite{nekoto2020participatory} established a participatory framework for low-resource machine
translation through the Masakhane community, emphasizing that effective MT
development for African languages requires both technical innovation and
community engagement. These studies share a common pattern: when the
target language is absent from the pretrained model, a proxy language
token is selected based on typological or genealogical similarity. The
choice is typically made once and not compared against alternatives.

These studies share a common pattern: when the target language is absent
from the pretrained model, a proxy language token is selected based on
typological or genealogical similarity. The choice is typically made once
and not compared against alternatives.

Alternative adaptation strategies exist. \cite{pfeiffer2020mad} proposed
MAD-X is an adapter-based framework that inserts language-specific adapter
modules rather than modifying embeddings. While adapters offer modular
extensibility, they add architectural complexity and require adapter
training infrastructure. Our averaging method requires no architecture
change beyond resizing the embedding matrix.

Embedding averaging has been explored in cross-lingual word embedding
alignment \cite{doval2018improving}, but to our knowledge, it has not been evaluated as a language-token initialization strategy for multilingual NMT models. The
key difference is that we average language identity embeddings, i.e., single
vectors that condition the entire encoder-decoder, rather than
word-level representations. Our work addresses the proxy selection gap
by proposing and evaluating this initialization on a genuinely unseen
language.

We will like to acknowledge that this work builds on the broader effort of the Masakhane community \cite{adelani2021masakhaner} to develop NLP resources and benchmarks
for African languages. Our corpus release for Limbum contributes to this
growing ecosystem.

\subsection{Limbum}

Limbum belongs to the Nkambe subgroup of Mbam-Nkam within the Grassfields
languages (Southern Bantoid; Glottolog: limb1268). It is spoken primarily
in the Nkambe and Ndu subdivisions of the Northwest Region of Cameroon. The
language uses a Latin-based orthography with Unicode combining diacritics to
mark three-level tones (high: acute accent, low: grave accent, mid:
unmarked). These diacritics are semantically contrastive: minimal pairs such
as \textit{bə̀} (``to come'') vs. \textit{bə́} (``to be'') differ only in tone marking.

No computational resources for Limbum existed before this work. The language is classified as "developing" (EGIDS level 5) by Ethnologue \cite{eberhard2020ethnologue}, indicating active use in daily life and some written materials, but no digital NLP presence.

\section{Method}
\subsection{Embedding Averaging for Unseen Languages}

Given a pretrained multilingual NMT model $\mathcal{M}$ with language token
set $\mathcal{L}$, and an unseen language $L \notin \mathcal{L}$, we add $L$
to $\mathcal{L}$ by the following procedure:

\begin{enumerate}
    \item \textbf{Select proxy languages.} Choose $k$ languages
    $\{P_1, \ldots, P_k\} \subset \mathcal{L}$ from the same language family
    as $L$.
    \item \textbf{Compute the averaged embedding.} Initialize the embedding of $L$ as:
    \begin{equation*}
        \mathbf{e}_L = \frac{1}{k} \sum_{i=1}^{k} \mathbf{e}_{P_i}
    \end{equation*}
    \item \textbf{Resize the model.} Add the new token to the tokenizer and expand the
    embedding matrix by one row. The new row is set to $\mathbf{e}_L$.
\end{enumerate}

\noindent The method makes one assumption: that related languages occupy a coherent
region of the pretrained embedding space so that their centroid lies within
that region rather than at an arbitrary point. This assumption is grounded.
In prior observations, multilingual models learn language representations
that cluster by family \cite{johnson2017google}, though we note it remains
an empirical question whether the centroid is a better starting point than
any individual proxy.

\noindent \textbf{Proxy selection for Limbum.} We select $k = 3$ Bantu languages present in NLLB-200 shown in Table \ref{tab:proxies}.

\begin{table}[t] % Floated to top per instructions
    \centering
    \small 
    \setlength{\tabcolsep}{4pt} % Tighten spacing to fit column
    \begin{tabular}{lllp{3.5cm}}
        \specialrule{1.5pt}{0pt}{0pt} % Thick top line
        \textbf{Proxy} & \textbf{ISO} & \textbf{Zone} & \textbf{Rationale} \\ 
        \specialrule{1pt}{0pt}{0pt}   % Thin line below header
        Swahili & \texttt{swh\_Ltn} & E.70 & Highest-resource Bantu in NLLB \\ 
        Luganda & \texttt{lug\_Ltn} & J.10 & Agglutinative; strong NLLB representation \\ 
        Lingala & \texttt{lin\_Ltn} & C.30 & Proximity to Grassfields zone \\ 
        \specialrule{1.5pt}{0pt}{0pt} % Thick bottom line
    \end{tabular}
    \caption{Bantu proxy languages selected for embedding initialization.}
    \label{tab:proxies}
\end{table}

\noindent We acknowledge a limitation: none of these three languages belongs to the
Grassfields Bantu subgroup that includes Limbum. They are selected based on Bantu family membership and availability in NLLB-200, not close
typological proximity. Whether closer relatives (if they were in the model)
would yield better initialization remains an open question.

\subsection{Fine-Tuning Protocol}
We fine-tune NLLB-200-distilled-600M (\texttt{facebook/nllb-200-distilled-600M}) 
using the Hugging Face \texttt{Seq2SeqTrainer}. For S2b, we apply a 
two-phase training schedule:

\begin{itemize}
    \item \textbf{Phase 1 (warm-up):} For the first 1,000 steps, we freeze all model parameters except the newly added \texttt{lmp\_Latn} embedding and the cross-attention parameters of the last decoder layer (keys, values, queries, and output projection). This allows the new embedding to adapt to the model's representation space before the full model begins to shift.

    \item \textbf{Phase 2 (full fine-tuning):} All parameters are unfrozen. We train with AdamW (learning rate $5 \times 10^{-5}$, linear warmup over 500 steps), maximum sequence length 128 tokens, effective batch size 16 (batch size 2 with gradient accumulation over 8 steps), and mixed precision (fp16). Training runs for a maximum of 5 epochs with early stopping on the development-set chrF2++.
\end{itemize}

\noindent The conservative learning rate and small effective batch size are designed to reduce catastrophic forgetting of NLLB-200's multilingual representations when adapting to a small corpus. Gradient checkpointing is enabled to fit within the Google Colab memory of a T4 GPU.

\noindent For S2a, the protocol is identical except that the Swahili token \texttt{`swh\_Latn`} is used as the source language identifier without any embedding modification, and no warm-up freeze is applied (since no new parameters are introduced). We note that this means the S2a–S2b comparison conflates embedding
initialization with training schedule; an S2a variant with the same freeze phase would isolate the initialization effect more cleanly. We chose not to add this variant because our primary claim is that averaging matches the
single proxy, not that it exceeds it; the confound works against S2b
(fewer trainable steps in Phase 1), making our equivalence finding
conservative.

\subsection{Baseline Models}

\noindent \textbf{S0 (zero-shot)}: We run NLLB-200 without fine-tuning, mapping Limbum
input to \texttt{swh\_Latn} and requesting English output (\texttt{eng\_Latn}). This
measures what the pretrained model produces for an unseen Bantu language.

\vspace{1em} % Adds a small gap between the two baselines

\noindent \textbf{S1 (Transformer from scratch)}: A standard transformer \cite{vaswani2017attention} with 6 encoder and 6 decoder layers, model dimension 256, 4 attention heads, feed-forward dimension 512, and dropout 0.3. We use Byte Pair Encoding \cite{sennrich2016neural} with a joint vocabulary of 8,000 merge operations. This baseline isolates the contribution of the in-domain data from any pretrained knowledge.

\section{Data}

\subsection{Corpus Creation}
The parallel corpus was compiled from two primary sources: the \textit{New Testament} in Limbum, published by the Bible Society of Cameroon, and the \textit{Limbum Dictionary} by Francis Ndi Wepngong. Both resources were chosen for their linguistic coverage and availability in bilingual format.

The parallel corpus for this study was compiled from two primary bilingual resources: the Limbum New Testament published by the Bible Society of Cameroon and the Limbum Dictionary authored by Francis Ndi Wepngong. The New Testament, provided in PDF format, is aligned verse by verse with the American Standard Version (ASV) in English. The ASV was selected because of its relatively modern and consistent vocabulary, which facilitates alignment and reduces stylistic noise compared to older English versions. The New Testament was provided in PDF format. A rule-based parser was developed to extract verses using pattern matching. Each verse in Limbum was aligned with its corresponding English verse from the \textit{American Standard Version} (ASV). The ASV was selected because of its simple, modern style and consistency in vocabulary, which reduces noise in translation training compared to older or more literary English versions.

In parallel, the dictionary was processed to extract lemma-gloss pairs, providing lexical coverage that complemented the sentence-level alignments derived from the Bible. This combination of verse-aligned text and lexicographic entries constitutes the first structured attempt to develop a Limbum–English parallel resource, representing the central novelty of this work. The Limbum dictionary was also processed with pattern matching to extract individual entries. Each lemma in Limbum was paired with its English glosses, and formatting artifacts, such as page headers, irregular spacing, or hyphenation, were removed. These dictionary pairs provided lexical coverage that complemented the sentence-aligned Bible corpus.

Following extraction, the texts were subjected to a series of cleaning and normalization steps. All limbum sentences were standardized to Unicode NFC to ensure a consistent representation of diacritics and accented forms. Rule-based parsers were developed to detect and extract verse identifiers in the Bible, while pattern-matching techniques isolated headwords and glosses in the dictionary. Both sources were stripped of page headers, irregular spacing, hyphenation, and other formatting artifacts. Sentence alignment was performed using verse identifiers for the scriptural text, while dictionary items were directly paired with their English glosses. Only entries with nonempty and alphabetic content on both sides were retained.
\subsection{Data Preprocessing}

\begin{figure}[t] % Floated to top
    \centering
    \includegraphics[width=\columnwidth]{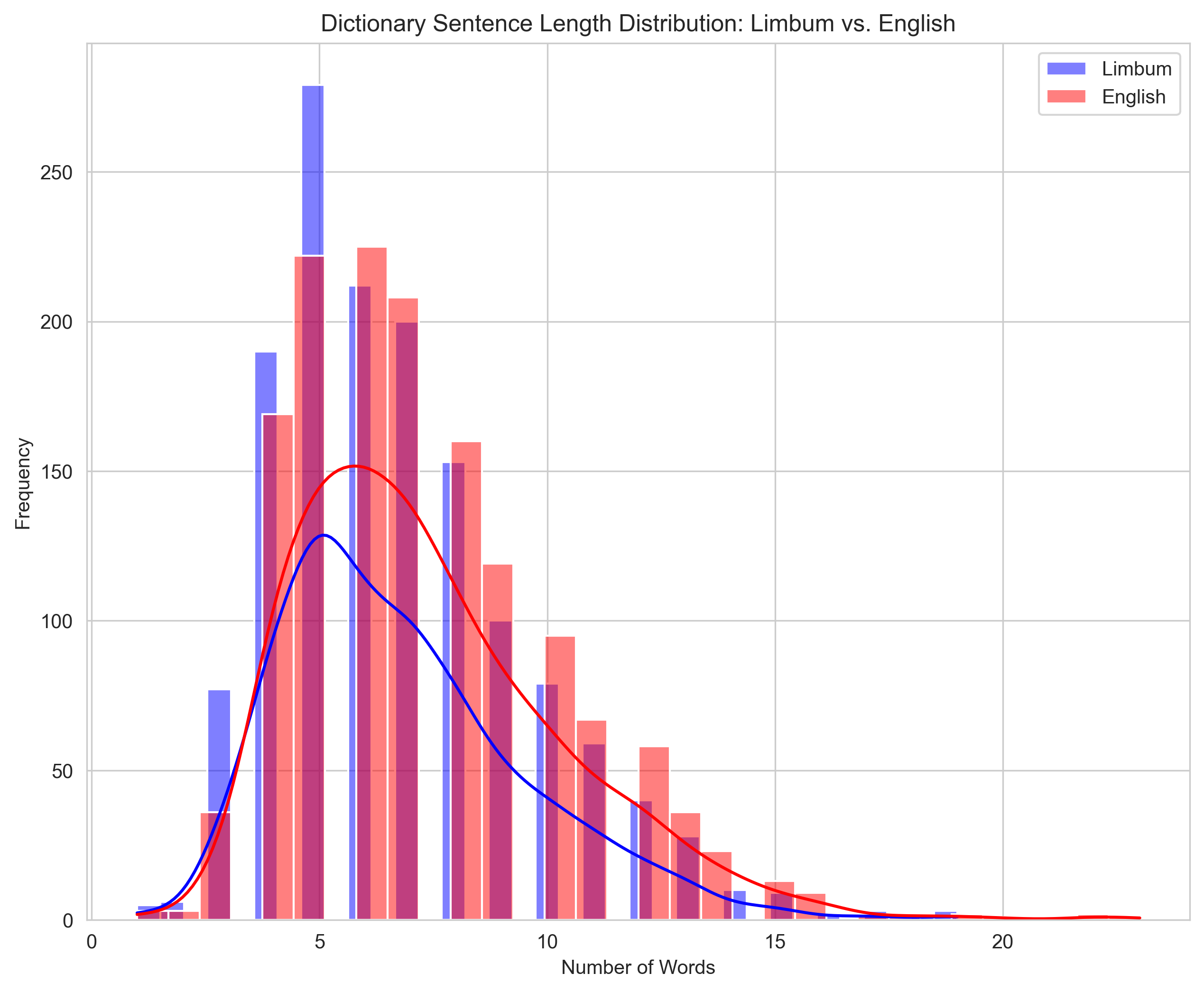}
    \caption{Sentence length distribution (number of words) for Limbum vs. English in the dictionary corpus.}
    \label{fig:dist_dict}
    \vspace{-10pt} % Reclaims a bit of vertical space
\end{figure}

\begin{figure}[t] % Floated to top
    \centering
    \includegraphics[width=\columnwidth]{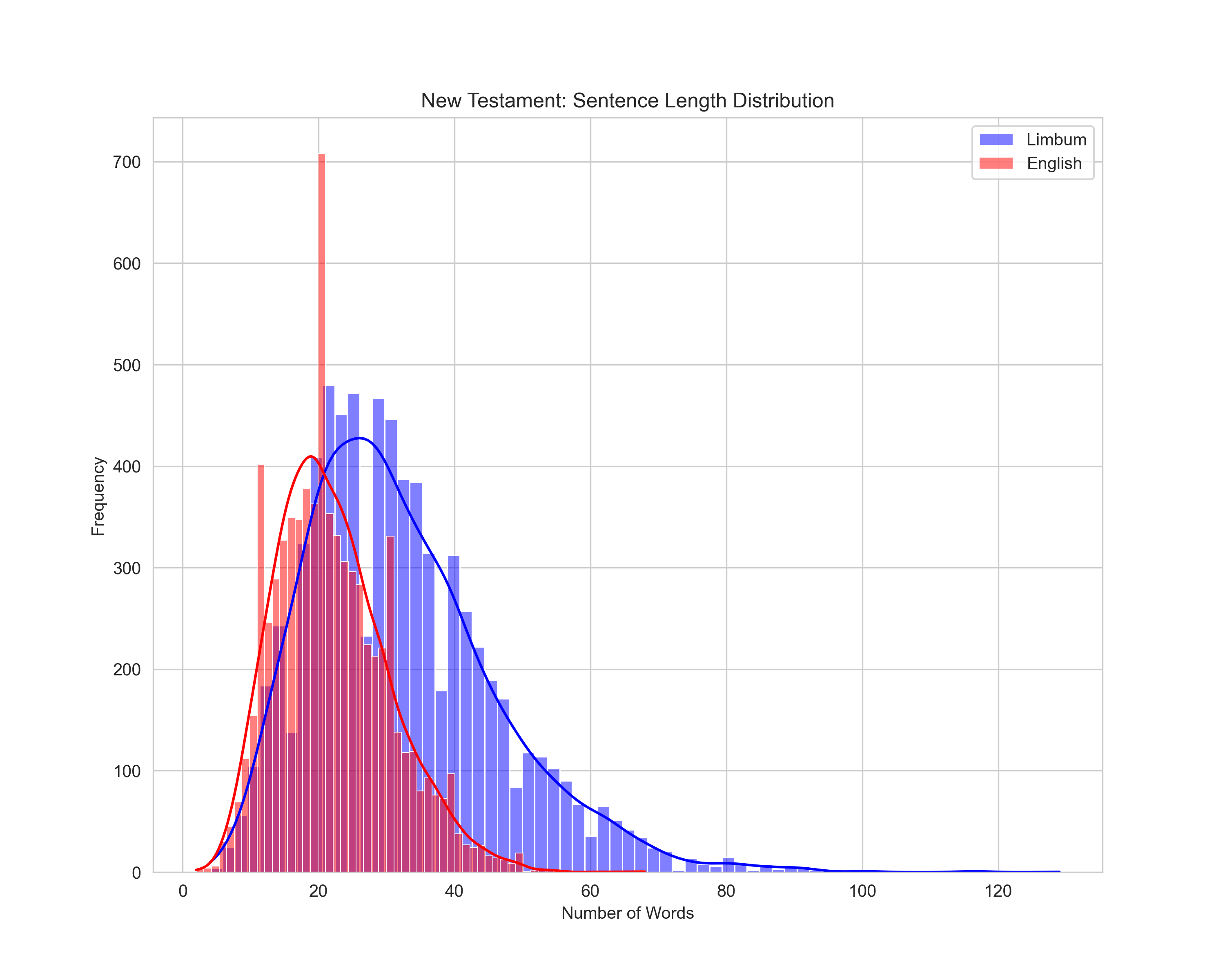}
    \caption{Sentence length distribution (number of words) for Limbum vs. English in the New Testament corpus.}
    \label{fig:dist_nt} % Changed label to be unique
    \vspace{-10pt} % Reclaims a bit of vertical space
\end{figure}

We compiled the first Limbum-English parallel corpus from two sources:

\begin{itemize}
    \item \textbf{Limbum New Testament:} We extract 7,380 verse-aligned sentence pairs from a published translation of the Limbum New Testament \cite{limbumNT2003} and its English equivalent (ASV). The Limbum New Testament is publicly distributed and used in churches across the Nkambe Plateau. Alignment is performed at the verse level, which provides natural sentence boundaries for Biblical text. After alignment, we remove exact duplicates and pairs where either side is empty or exceeds 200 tokens. Sentence length distribution for this corpus in Figure \ref{fig:dist_nt}.

    \item \textbf{Bilingual dictionary:} We digitize 1,457 entry pairs from a published Limbum-English dictionary compiled by SIL linguists working with Wimbum community members. Dictionary entries tend to be shorter (mean 7.5 Limbum tokens vs. 32.3 for NT) and lexically diverse, covering everyday vocabulary outside the religious domain. Sentence length distribution for this corpus in Figure \ref{fig:dist_dict}.
\end{itemize}

The combined corpus contains \textbf{8,837 parallel pairs} after cleaning and preprocessing. All text undergoes Unicode NFC normalization to ensure consistent representation of combining diacritics. We release the corpus under a \mbox{CC-BY-SA 4.0} license. We acknowledge that the two sources were produced for the Wimbum community, and we were able to get in touch with one of the main authors to get approval. Our computational use is intended to benefit speakers of the language by enabling machine translation research.

\subsubsection{Data Statistics and Splits}
We partition the corpus into train (80\%), development (10\%), and test (10\%)
splits using stratified sampling by source (NT or dictionary), so that both
domains appear proportionally in all splits. Values shown in Table \ref{tab:splits}.

\begin{table}[t] % Floated to top per instructions
\centering
\begin{tabular}{lrrr}
\specialrule{1.5pt}{0pt}{0pt} % Thick top line (1.5pt)
\textbf{Split} & \multicolumn{1}{r}{\textbf{NT}} & \multicolumn{1}{r}{\textbf{Dictionary}} & \multicolumn{1}{r}{\textbf{Total}} \\ 
\specialrule{1pt}{0pt}{0pt}   % Thin line below header (1pt)
Train & 5,904 & 1,166 & 7,070 \\
Dev   & 738   & 146   & 884   \\
Test  & 739   & 146   & 885   \\ 
\specialrule{1.5pt}{0pt}{0pt} % Thick bottom line (1.5pt)
\end{tabular}
\caption{Corpus splits stratified by source domain.}
\label{tab:splits}
\end{table}

Limbum sentences average 32.3 words (NT) and 7.5 words (dictionary).
English sentences average 22.1 words (NT) and 5.2 words (dictionary). The
Limbum vocabulary contains 6,734 unique word types compared to 5,198 for
English, reflecting the agglutinative morphology of the language: Limbum
encodes grammatical information through prefixes and suffixes, producing
more surface forms per lemma.

\section{Experiments and Results}

\subsection{Evaluation Metrics}

We report two automatic metrics. \textbf{chrF2++} \cite{popovic2017chrf++} is our primary
metric: a character-level n-gram F-score with $\beta = 2$ recall weighting.
It is robust at small corpus sizes and captures partial morpheme matches,
making it well-suited for agglutinative languages. \textbf{SacreBLEU} \cite{post2018call} is reported as a secondary metric for comparability with prior NMT
work (signature: \texttt{nrefs:1|case: lc|eff:no|tok:13a|smooth:exp}).
We report 95\% bootstrap confidence intervals (1,000 resampling iterations)
for chrF2++ to assess whether observed differences are statistically significant
and reliable, given our test set of 885 pairs.

\subsection{Main Results}

Table~\ref{tab:results} presents results on the combined test set. Three observations stand out.

\begin{table}[t] % Floated to top
\centering
\small 
\setlength{\tabcolsep}{3pt} 
\begin{tabular}{l p{2.8cm} r r r}
\specialrule{1.5pt}{0pt}{0pt} % Thick top line
\textbf{Sys} & \textbf{Description} & \multicolumn{1}{r}{\textbf{chrF}} & \multicolumn{1}{r}{\textbf{BLEU}} & \multicolumn{1}{r}{\textbf{95\% CI}} \\ 
\specialrule{1pt}{0pt}{0pt}   % Thin line below header
S0  & NLLB zero-shot & 12.52 & 0.66 & [---, 13.02] \\
S1  & Transf. (scratch) & 14.47 & 1.32 & [---, 14.86] \\
S2a & NLLB + \texttt{swh\_Latn} & 47.27 & 30.56 & [---, 48.52] \\
S2b & NLLB + \texttt{lmp\_Latn} & 46.70 & 29.88 & [45.3, 47.9] \\ 
\specialrule{1.5pt}{0pt}{0pt} % Thick bottom line
\end{tabular}
\caption{Main results ($N = 885$)}
\label{tab:results}
\end{table}

\textbf{Multilingual transfer is the dominant factor.} The gap between S1 and S2a is 32.8 chrF2++ points ($14.47 \rightarrow 47.27$). By contrast, the gap between S2a and S2b is 0.57 points, and the confidence intervals overlap substantially. Pretrained multilingual knowledge accounts for the vast majority of the performance gain; the initialization strategy is a secondary consideration.

\textbf{Averaged initialization matches the single proxy.} S2b achieves chrF2++ of 46.70 compared to 47.27 for S2a, a difference of 0.57 points. The S2b 95\% confidence interval $[45.29, 47.90]$ overlaps with the S2a upper bound of 48.52. We note that overlapping intervals do not formally establish equivalence; however, the observed difference (0.57) is an order of magnitude smaller than the S1--S2a gap (32.8) and falls well within the range attributable to test-set variance at $N=885$. This supports our claim that researchers need not search for the "best" proxy language: averaging across multiple related languages produces comparable results.

\textbf{Zero-shot transfer is near-zero.} S0 produces a chrF2++ of only 12.52, confirming that NLLB-200 has no usable knowledge of Limbum without fine-tuning. The model appears to generate Swahili-influenced output that shares some character overlap with Limbum (both being Bantu) but carries no translation utility.

\textbf{Tonal diacritics remain unsolved.} Diacritic accuracy, the proportion of output tokens that exactly reproduce reference diacritics, is 0.0\% across all models. No model produces correct tonal markings on any test token. This indicates a systemic failure in tone preservation that chrF2++ and BLEU only partially penalize, since a hypothesis with correct lexemes but stripped diacritics still receives substantial character overlap credit.

\subsection{Domain Stratification}

Table~\ref{tab:domain_results} stratified results by source domain.

\begin{table}[t] % Floated to top
\centering
\footnotesize
\setlength{\tabcolsep}{4pt} % Slightly more space since vertical lines are gone
\begin{tabular}{l rr rr}
\specialrule{1.5pt}{0pt}{0pt} % Thick top line (1.5pt)
\textbf{Sys} & \multicolumn{2}{c}{\textbf{NT ($N=739$)}} & \multicolumn{2}{c}{\textbf{Dict ($N=146$)}} \\
\cmidrule(lr){2-3} \cmidrule(lr){4-5} % Standard LaTeX lines for grouped headers
 & \multicolumn{1}{r}{\textbf{chrF}} & \multicolumn{1}{r}{\textbf{BLEU}} & \multicolumn{1}{r}{\textbf{chrF}} & \multicolumn{1}{r}{\textbf{BLEU}} \\ 
\specialrule{1pt}{0pt}{0pt}   % Thin line below header (1pt)
S0  & 12.62 & 0.67 & 10.54 & 0.16 \\
S1  & 15.56 & 1.51 & 6.36  & 0.34 \\
S2a & 48.65 & 31.63 & 24.43 & 9.53 \\
S2b & 47.95 & 30.87 & 25.48 & 9.68 \\ 
\specialrule{1.5pt}{0pt}{0pt} % Thick bottom line (1.5pt)
\end{tabular}
\caption{Results stratified by domain.}
\label{tab:domain_results}
\end{table}

All systems perform better on NT pairs than dictionary pairs, which is expected given that NT constitutes 84\% of the training data. The difference is most pronounced for S1, where dictionary chrF2++ (6.36) falls well below NT (15.56), suggesting the from-scratch model memorizes domain-specific patterns rather than learning generalizable translations.

On dictionary pairs, S2b scores slightly higher than S2a (25.48 vs. 24.43), while on NT pairs S2a leads by a similar margin (48.65 vs. 47.95). Neither difference is statistically testable at conventional confidence levels given the small domain-specific subsets ($N=146$ for dictionary, $N=739$ for NT). We therefore do not conclude domain-specific advantages of either initialization strategy from these results.

\subsection{Data Efficiency}

To assess how translation quality scales with training data, we train S2b
on 10\%, 25\%, 50\%, and 75\% subsets of the training data (sampled with
stratified proportions from both NT and dictionary sources) and evaluate
on the full test set. Table~\ref{tab:learning_curve} reports the results.

\begin{table}[t] % Floated to top
\centering
\small
\setlength{\tabcolsep}{3.5pt}
\begin{tabular}{r r r r r}
\specialrule{1.5pt}{0pt}{0pt} % Thick top line
\multicolumn{1}{r}{\textbf{Frac.}} & \multicolumn{1}{r}{\textbf{N (train)}} & \multicolumn{1}{r}{\textbf{chrF2++}} & \multicolumn{1}{r}{\textbf{BLEU}} & \multicolumn{1}{r}{\textbf{95\% CI}} \\ 
\specialrule{1pt}{0pt}{0pt}   % Thin line below header
10\%  & 707   & 20.43 & 4.03  & [19.80, 21.08] \\
25\%  & 1,768  & 32.12 & 14.27 & [31.12, 33.07] \\
50\%  & 3,535  & 39.26 & 21.67 & [37.96, 40.46] \\
75\%  & 5,303  & 44.32 & 27.86 & [43.03, 45.54] \\
100\% & 7,070  & 46.70 & 29.88 & [45.29, 47.90] \\ 
\specialrule{1.5pt}{0pt}{0pt} % Thick bottom line
\end{tabular}
\caption{S2b learning curve across training data fractions.}
\label{tab:learning_curve}
\end{table}

\noindent The result reveals a clear pattern of diminishing returns. The
largest single gain occurs between 10\% and 25\% of the data ($+11.7$
chrF2++ points), where the model transitions from near-baseline performance
to meaningful translation. Subsequent doublings yield progressively smaller
gains: $+7.1$ points from 25\% to 50\%, $+5.1$ from 50\% to 75\%, and $+2.4$ from
75\% to 100\%. At just 25\% of the training data (1,768 pairs), S2b already
achieves chrF2++ of 32.12, nearly 18 points above the from-scratch
baseline (S1 = 14.47), demonstrating that multilingual transfer provides
substantial benefit even with very limited fine-tuning data. At 75\%
(5,303 pairs), The model reaches within 2.4 points of full-data
performance (44.32 vs. 46.70), suggesting that practitioners with limited
Resources can achieve most of the translation quality with substantially
less annotation effort.

\subsection{NT-Only Ablation}

To measure the contribution of the bilingual dictionary pairs, we train
S2b on New Testament data only (5,904 training pairs, excluding all 1,166
dictionary pairs) and evaluate on the full test set. Table~\ref{tab:ablation} reports
results alongside the full-corpus S2b for comparison.

\begin{table}[t] % Floated to top
\centering
\footnotesize
\setlength{\tabcolsep}{4pt} 
\begin{tabular}{l r r r r r}
\specialrule{1.5pt}{0pt}{0pt} % Thick top line
\textbf{Training Data} & \multicolumn{1}{r}{\textbf{N}} & \multicolumn{2}{c}{\textbf{Full Test}} & \multicolumn{1}{r}{\textbf{NT}} & \multicolumn{1}{r}{\textbf{Dict}} \\
\cmidrule(lr){3-4}
& & \multicolumn{1}{r}{\textbf{chrF}} & \multicolumn{1}{r}{\textbf{BLEU}} & \multicolumn{1}{r}{\textbf{chrF}} & \multicolumn{1}{r}{\textbf{chrF}} \\ 
\specialrule{1pt}{0pt}{0pt}   % Thin line below header
NT only   & 5,904 & 45.58 & 29.31 & 47.44 & 15.30 \\
NT + Dict & 7,070 & 46.70 & 29.88 & 47.95 & 25.48 \\ 
\specialrule{1.5pt}{0pt}{0pt} % Thick bottom line
\end{tabular}
\caption{Effect of dictionary data on S2b performance.}
\label{tab:ablation}
\end{table}

\noindent Removing the dictionary pairs has a modest effect on overall performance
(45.58 vs. 46.70, a drop of 1.12 chrF2++ points) and a negligible effect
on NT-domain test items (47.44 vs. 47.95). However, the impact on
dictionary-domain test items is dramatic: chrF2++ drops from 25.48 to
15.30 --- a loss of 10.18 points, bringing performance close to the
from-scratch baseline. This confirms that the dictionary pairs are
essential for out-of-domain lexical coverage. The short, diverse dictionary
entries expose the model to everyday vocabulary absent from the New
Testament, and without them, the system fails to translate non-religious
content. For researchers assembling corpora for new languages, this
result argues strongly for supplementing any primary text source with
lexical resources, even if the lexicon is small (1,166 training pairs in
our case, just 16\% of the total corpus).

\section{Discussion}

\subsection{Why Does Averaging Work?}

The embedding averaging method rests on the observation that NLLB-200's
language embeddings are not randomly distributed. Languages from the same
family tend to cluster in the embedding space \cite{johnson2017google}. The
centroid of three Bantu language embeddings, therefore, lies in a region of
the space associated with Bantu-family features: agglutinative morphology,
subject-verb-object word order, and noun class systems. This provides a
reasonable initialization for gradient descent, even though the centroid
does not correspond to any specific language.

Notably, our three proxy languages (Swahili, Luganda, and Lingala) are not
close relatives of Limbum within the Bantu tree. They span Eastern and
Northwestern zones, while Limbum belongs to the Grassfields subgroup.
That the method works despite this distance suggests the embedding captures
broad family-level features rather than requiring fine-grained typological
matching. This is a promising property for practical deployment, since many
unseen languages will lack close relatives in the model.

A natural question is whether any initialization, including random, 
would converge to a similar quality given 7,070 fine-tuning pairs. We did
not include a random-initialization baseline, which limits our ability to
isolate the specific benefit of family-informed averaging. However, prior
work on transfer learning suggests that initialization affects convergence
speed and stability even when final performance converges \cite{zoph2016transfer,neubig2018rapid}. The practical advantage of our method is not
necessarily final-quality superiority over random initialization but rather that it provides a principled, robust default that avoids the risk of pathological starting points. This property becomes more valuable as the corpus size decreases. We leave systematic comparison with random and single-language initializations across multiple languages to future work.

The learning curve analysis in table \ref{tab:learning_curve} provides additional context: with only 25\% of the training data (1,768 pairs), S2b already reaches a chrF2++ of 32.12, more than double the from-scratch baseline, and at 75\%, it
achieves 44.32, within 2.4 points of the full-data result. This confirms that the method benefits substantially from multilingual transfer, even at very small corpus sizes, reinforcing its practical value for researchers who may have limited parallel data available.

\subsection{Practical Implications}

The embedding averaging method has two practical advantages for researchers
working with unseen African languages.

First, it eliminates the problem of proxy selection. A researcher adding a new Bantu language to NLLB-200 need not determine which single language is the
best proxy. They can average over any available Bantu tokens in the model.
Our results suggest the performance cost of this convenience is negligible.

Second, the entire pipeline corpus preparation, model fine-tuning, and
evaluation runs on a single T4 GPU (15 GB VRAM) available
through Google Colab Pro. No proprietary infrastructure is required. This is
relevant for the many African language research groups that operate without
institutional computing budgets. We release all code and the training pipeline
to enable direct replication.

Potential downstream applications would include translation tools for educational materials for Limbum-medium primary schools and digital
resources for the Limbum-speaking diaspora. We stress that deployment
would require community validation by native speakers — a step beyond
the scope of this study, but essential before any real-world use.

\subsection{The Tonal Diacritic Problem}

The most striking failure across all four systems is the complete absence of correct tonal diacritics. Since Limbum uses tone contrastively, stripped diacritics produce genuinely ambiguous output; a word like \textit{bə} without tone marking could mean ` "to come," ` "to be," or other meanings depending on context.

We hypothesize three contributing factors to this failure:

\begin{itemize}
    \item \textbf{Tokenizer representation.} NLLB-200's SentencePiece tokenizer treats combining diacritics (Unicode combining acute \texttt{U+0301}, combining grave \texttt{U+0300}) as separate tokens or merges them unpredictably. During tokenization, a word like \textit{bə̀} may be split into subword units that separate the base character from its diacritic, making it difficult for the model to learn the association between lexical identity and tone marking.

    \item \textbf{Training data sparsity.} Our training corpus contains 7,070 pairs, but the number of unique tonal patterns is much larger. Each Limbum word can carry one of three tones on each syllable, creating a combinatorial space that is poorly sampled at this data size.

    \item \textbf{Evaluation direction.} Our primary evaluation is Limbum$\rightarrow$English, where tonal diacritics appear only on the source side. The model need not \textit{produce} diacritics to achieve a high chrF2++ on English output. The diacritic failure would be more visible and consequential in the reverse direction (English$\rightarrow$Limbum), which we did not evaluate.
\end{itemize}

This finding has implications beyond Limbum. Many African languages use tonal diacritics that are contrastive, including Yoruba, Igbo, and numerous Bantu languages. NMT systems that strip or ignore tone markings may produce output that is superficially fluent but semantically ambiguous. We argue that diacritic-aware evaluation metrics, beyond standard chrF2++ and BLEU, are needed for tonal languages.

\subsection{Limitations}

We identify five limitations of this study.

\begin{enumerate}
    \item \textbf{Single language pair.} We evaluate in Limbum-English only. The method needs replication on other unseen languages and language families before generalization claims can be made.

    \item \textbf{Fixed proxy set.} We use $k = 3$ proxy languages without exploring the effect of different values of $k$ or different proxy combinations. Weighted averaging by typological distance is an unexplored alternative.

    \item \textbf{Small test set.} With 885 test pairs, our confidence intervals are relatively wide ($\pm 1.3$chrF2++), limiting statistical power for detecting small differences between S2a and S2b.

    \item \textbf{Domain coverage}. The corpus is drawn from religious texts, which may not represent conversational Limbum. The use of the dictionary widens the scope to include a variety of words that mitigate this. However, additional context would be required to extend it fully to other domains on which performance remains unknown.

    \item \textbf{Tonal diacritics.} All systems produced zero correct diacritics in our evaluation, indicating a systemic failure in tone-marking that our current metrics (chrF2++, BLEU) only partially capture. Dedicated evaluation of diacritic fidelity is needed in future work.
\end{enumerate}

\section{Conclusion}

We have presented the first parallel corpus for Limbum–English (8,837
pairs from two domains) and demonstrated that fine-tuning NLLB-200 on
this corpus yields chrF2++ $\approx47$ for Limbum-to-English translation, a
32-point improvement over a transformer trained from scratch. This
confirms that multilingual transfer is the dominant factor for extremely
low-resource Bantu translation, consistent with findings across other
African language pairs (Adelani et al., 2022).

Our embedding averaging method, which involves initializing a new language token as the
centroid of three related Bantu language embeddings,  matches single-proxy
fine-tuning in practice (46.7 vs. 47.3 chrF2++), providing a convenient
default that eliminates the proxy selection problem. The learning curve analysis shows that 25\% of the training data (1,768 pairs) already yields
a chrF2++ of 32.12, and the NT-only ablation demonstrates that even a small supplementary dictionary (16\% of the corpus) is critical for
out-of-domain coverage.

The complete failure of all models on tonal diacritics is a finding with
implications beyond Limbum. For tonal African languages, standard NMT
pipelines may produce output that is fluent but tonally ambiguous.  The development of diacritic-aware evaluation metrics and dedicated modeling strategies for tone preservation is a potential research gap for exploration.

Future work should prioritize three directions: (1) evaluation of the
English→Limbum direction, which is more directly relevant for the community
deployment (translating health and educational materials into Limbum);
(2) investigation of tone-aware tokenization and decoding strategies;
and (3) participatory evaluation with Wimbum community members to
establish which translation capabilities are most needed and to validate
output quality.

We release the 8,837-pair corpus and all code at
\url{https://anonymous.4open.science/r/limbum_translation-E4B0/} and encourage replication
of this approach on other unseen languages within and beyond the Bantu
family. Future work should evaluate the reverse direction (English$\rightarrow$Limbum),
which is more relevant for community deployment scenarios such as translating health or educational materials into Limbum, and investigate dedicated strategies for tonal diacritic preservation.

% \subsubsection{Order of Sections}
% Sections should be arranged in the following order:
% \begin{enumerate}
%     \item Main content sections (numbered)
%     \item Appendices (optional, numbered using capital letters)
%     \item Ethical statement (optional, unnumbered)
%     \item Acknowledgements (optional, unnumbered)
%     \item Contribution statement (optional, unnumbered)
%     \item References (required, unnumbered)
% \end{enumerate}

\bibliographystyle{named}
\bibliography{ijcai26}

\end{document}